\documentclass[letterpaper]{article}
\usepackage{aaai}
\usepackage{times}
\usepackage{helvet}
\usepackage{courier}
\usepackage{soul}
\usepackage{xcolor}

\usepackage{setspace}
\usepackage[multiple]{footmisc}
\usepackage{bigfoot}
\usepackage{amsmath}
\usepackage{array}
\usepackage{float}
\usepackage{algorithm}
\usepackage{mathtools} 
\usepackage{subcaption}
\usepackage{algpseudocode}
\usepackage{amsthm}
\algnewcommand{\LineComment}[1]{\State \(\triangleright\) #1}

\DeclareNewFootnote{AAffil}[arabic]
\DeclareNewFootnote{ANote}[fnsymbol]
\usepackage{tcolorbox}
\usepackage[skip=1ex]{caption} 
\usepackage{tabularx}
\usepackage{array}
\usepackage{amsmath}
\usepackage{booktabs} 
\usepackage{lipsum} 
\usepackage{bigstrut}
\usepackage{booktabs}
\usepackage[hyphens]{url}
\usepackage{hyperref}
\usepackage[hyphenbreaks]{breakurl}

\algdef{SE}[SUBALG]{Indent}{EndIndent}{}{\algorithmicend\ }%
\algtext*{Indent}
\algtext*{EndIndent}

\begin{document}

\title{Testing the Limits of Unified Sequence to Sequence LLM Pretraining on Diverse Table Data Tasks}
\author{Soumajyoti Sarkar, Leonard Lausen \\
Amazon Web Services AI, US\\
\{soumajs, lausen\}@amazon.com\\
}
\maketitle

\begin{abstract}
Tables stored in databases and tables which are present in web pages and articles account for a large part of semi-structured data that is available on the internet. It then becomes pertinent to develop a modeling approach with large language models (LLMs) that can be used to solve diverse table tasks such as semantic parsing, question answering as well as classification problems. Traditionally, there existed separate models specialized for each task individually. It raises the question of how far can we go to build a unified model that works well on some table tasks without significant degradation on others. To that end, we attempt at creating a shared modeling approach in the pretraining stage with encoder-decoder style LLMs that can cater to diverse tasks. We evaluate our approach that continually pretrains and finetunes different model families of T5 with data from tables and surrounding context, on these downstream tasks at different model scales. Through multiple ablation studies, we observe that our pretraining with self-supervised objectives can significantly boost the performance of the models on these tasks. As an example of one improvement, we observe that the instruction finetuned public models which come specialized on text question answering (QA) and have been trained on table data still have room for improvement when it comes to table specific QA. Our work is the first attempt at studying the advantages of a unified approach to table specific pretraining when scaled from 770M to 11B sequence to sequence models while also comparing the instruction finetuned variants of the models.


\end{abstract}

\section{Introduction}
In recent years, there has been an emerging and quite substantial interest in specializing these LLMs \cite{chowdhery2022palm,wang2021codet5} on tasks with semi-structured data, for example knowledge graphs, spreadsheets, relational databases and tables which are of interest in this paper. Notably, there have been three streams of work that have been gaining traction in the realm of LLM pretraining for table specific tasks: 

\begin{enumerate}
    \item The first group of studies focus on predicting labels (essentially one column of a table) for classification and regression scenarios, using row values and column schema (from the other columns) as input \cite{huang2020tabtransformer,arik2021tabnet,somepalli2021saint,gorishniy2021revisiting};  \cite{grinsztajn2022why,wang2022transtab,Du2022,wydmanski2023hypertab}. They use gradient descent-based end-to-end learning with task-specific model pretraining and fine-tuning.
    \item The second group of studies aim to specialize LLMs on table data to retrieve task-agnostic table/column/row representations for different downstream table understanding tasks. Drawing inspiration from models like BERT~\cite{devlin-etal-2019-bert}, these studies ~\cite{herzig-etal-2020-tapas,yin20acl,deng2020turl,iida-etal-2021-tabbie} serialize tables to a sequence of tokens and train them on textual self-supervised objectives.
    \item Yet, there is a third group of studies that aim to use the generative nature of language models for data-to-text generation tasks where the nature of \textit{data} entails some form of table and/or related text and the output \textit{text} generally corresponds to text from the table or an output like SQL \cite{andrejczuk2022table,shi2022generation,parikh2020totto,liu2021tapex}.
\end{enumerate}

Putting the above into the landscape of work that has been popular, the goal of our study is to conduct self-supervised pre-training with table specific data along with text related to the tables as input, followed by pre-finetuning for improving generalization to diverse tasks. While it is clear that there has been disparate studies with tabular data, some aiming at  Question Answering (QA) tasks, some at getting row/column/table representations for semantic search and some at more complex reasoning tasks, what has not been done so far is to meaningfully conduct a methodological study of \textit{large} scale pretraining with table data that can be \textit{either} finetuned for specific downstream table tasks \textit{or} help in scenarios where there is scarcity of data for a specific task. Specifically, the last group of studies in our above categorization focus on pretraining that can equip the models for end-end generation abilities suitable for table semantic parsing, table-based QA, table summarization among other tasks. However, pretraining with specialized objectives in the absence of large scale data and scale of models, remain key to transfer learning over diverse downstream tasks. We revisit several questions around table based LLM pretraining that have been studied before, albeit on a much smaller scale than the data used in this study. We attempt at answering the following:

\begin{itemize}
    \item \textbf{RQ1}: Can an intermediate pretraining stage with large scale data and multiple self-supervised text-text objectives transfer well to diverse downstream table tasks of both generation and encoding and can it improve performance on them compared to the text focused base models that we start with?
    \item \textbf{RQ2}: Can this intermediate pretraining along with added pre-finetuning help in improving the generalization capabilities of the models in situations where the downstream tasks have scarce training data?
    \item \textbf{RQ3}: Do we see improvements on these tasks from the table specific pretraining as we scale the model sizes while keeping the compute budget fixed in pretraining \cite{hoffmann2022training}?
\end{itemize}


Note that throughout this paper, although we measure the performance of our models with a few state-of-the-art baselines in controlled settings, we focus less on the nuances of any specific downstream task and instead measure the trade-offs between the gains and costs that come along with the unified model. In the literature of table based QA and reasoning, all narratives on improving performance on tasks involving  semi-structured  data has been either on unifying them through a knowledge grounded framework as the work conducted in UnifiedSKG \cite{xie2022unifiedskg} or through carefully selected and annotated  datasets in finetuning that work well on desired downstream tasks \cite{neeraja2021incorporating}. As it goes, the  data used for pretraining in these studies is miniscule compared to the pretraining regime of LLMs which are trained on trillion tokens \cite{hoffmann2022training}. We focus our efforts primarily in two specific areas: (1) pretraining encoder-decoder models in an attempt to improve on question answering and/or text/SQL (related to table) generation tasks, (2) extracting table or column level representations that can be used for classification tasks. Through our experiments, we observe the following:

\begin{itemize}
    \item Our models trained with self-supervised objectives starting from public T5 model checkpoints and continually trained on table task data achieves the best results and (near) state-of-the-art results on several downstream table generation tasks.
    \item Our models even outperform existing encoder models of similar sizes and specialized to tabular tasks on some of the popular table classification tasks.
    \item Our results show that we also achieve improvements when scaling to 11B even when our pretraining method keeps the compute budget near constant for the 11B compared to the 3B models. However, we do find that the gains are diminishing at larger model sizes when the size of the table related pretraining data is kept similar to the samller model training regime.
\end{itemize}

\section{Unified Language Model Pretraining} \label{sec:unified}
Before we get into the technical details of the pretraining and prefinetuning regime adopted in our work, we lay out some fundamental use-cases which primarily motivate the need to consolidate the pretraining phase of several tabular tasks (more details on these can be found in Section 1 of the Appendix). The fragmented nature of tabular data tasks can be roughly categorized into the following: (1) Generating SQL-like expressions with natural language queries \cite{sun2018semantic,lin2020bridging}, (2) Question Answering on tables \cite{herzig-etal-2020-tapas,chen2020open}, (3) Fetching representations of table elements for classification \cite{yin2020tabert,fan2022semantics,wang2021tuta} and, (4) Data to Text Generation \cite{chen2020logical,pietruszka2022stable}. In this paper, we evaluate our pretrained models on the first three use-cases with appropriate downstream tasks. The main contribution of our paper has been to revisit how individual language models had been trained for these use cases in silos and then design a unified pretraining stage so that a single model can transfer well to these tasks. As mentioned in the work done in UnifiedSKG \cite{xie2022unifiedskg}, since these tasks leverage  structured  knowledge  to  complete  user requests, the challenge lies in the heterogeneous nature of the  inputs  and the  outputs.  They  have been studied separately by different communities. To handle that challenge, we follow the paradigm used in UnifiedSKG (which only contains a single/multi-task finetuning stage), albeit we add a pretraining stage and propose \textit{\underline{Uni}}fied \textit{\underline{Tab}}ular \textit{\underline{P}}re\textit{\underline{t}}raining - \textbf{UniTabPT}. We follow the \texttt{pretraining $\rightarrow$ multi-task prefinetuning $\rightarrow$ finetuning} paradigm, however we run several ablation studies to show how our method generalizes better to downstream tasks than the paradigm in UnifiedSKG and previous task specific baselines.

\begin{table*}[!t]
\centering
{\small
\begin{tabular}{c|c|c}  
\hline
\textbf{UniTabPT Encoder Input} & \textbf{UniTabPT Decoder Input} & \textbf{UniTabPT Decoder Output}\\  \hline
\parbox[t]{8cm}{ \textcolor{blue}{$<$context$>$} \textcolor{red}{$<$text\_NL$>$} text half 1 \textcolor{gray}{$<$header$>$} col name 1 $|$ col name 2 $|$ $\ldots$ \textcolor{green}{$<$row$>$} 0 val 1 $|$ val 2 $|$ $\ldots$ \textcolor{green}{$<$row$>$} 1 val 1 $|$ val 2 $|$ \ldots \\ 
}    & \parbox[t]{4cm}{\textcolor{orange}{$<$NL\_completion$>$} \textcolor{red}{$<$text\_NL$>$} text half 2 \\
}   & \textcolor{red}{$<$text\_NL$>$} text half 2\\ \hline
\parbox[t]{8cm}{ \textcolor{blue}{$<$context$>$} \textcolor{red}{$<$text\_SQL$>$} sql half 1 \textcolor{gray}{$<$header$>$} col name 1 $|$ col name 2 $|$ $\ldots$ \textcolor{green}{$<$row$>$} 0 val 1  $|$ val 2  $|$$\ldots$ \textcolor{green}{$<$row$>$} 1 val 1 $|$ val 2$|$ \ldots \\ 
}    & \parbox[t]{4cm}{\textcolor{orange}{$<$SQL\_completion$>$} \textcolor{red}{$<$text\_SQL$>$} sql half 2 \\
}   & \textcolor{red}{$<$text\_SQL$>$} sql half 2\\ \hline
\parbox[t]{8cm}{ \textcolor{blue}{$<$context$>$}\textcolor{brown}{$<$missing\_context$>$}  \textcolor{gray}{$<$header$>$} col name 1 $|$ col name 2 $|$ $\ldots$ \textcolor{green}{$<$row$>$} 0 val 1 $|$ val 2 
 $|$ $\ldots$ \textcolor{green}{$<$row$>$} 1 val 1 $|$ val 2  $|$ \ldots \\ 
}    & \parbox[t]{4cm}{\textcolor{orange}{$<$NL\_generation$>$} \textcolor{red}{$<$text\_NL$>$} text \\
}    & \textcolor{red}{$<$text\_NL$>$} text \\ \hline
\parbox[t]{8cm}{ \textcolor{blue}{$<$context$>$} \textcolor{red}{$<$text\_NL$>$} t1 \textcolor{magenta}{$<$sentinel 1$>$} t3 \textcolor{gray}{$<$header$>$} col name 1 $|$ \textcolor{magenta}{$<$sentinel 2$>$}$|$ $\ldots$ \textcolor{green}{$<$row$>$} 0 val 1 $|$ \textcolor{magenta}{$<$sentinel 3$>$} t6 $|$ $\ldots$ \textcolor{green}{$<$row$>$} 1 val 1 $|$ val 2 $|$ \ldots \\ 
}    & \parbox[t]{4cm}{\textcolor{orange}{$<$denoising$>$} \textcolor{magenta}{$<$sentinel 1$>$} t2 \textcolor{magenta}{$<$sentinel 2$>$} t4 \textcolor{magenta}{$<$sentinel 3$>$} t5 \\
}  & \parbox[t]{4cm}{\textcolor{magenta}{$<$sentinel 1$>$} t2 \textcolor{magenta}{$<$sentinel 2$>$} t4 \textcolor{magenta}{$<$sentinel 3$>$} t5} \\ \hline
\end{tabular}
}
\vspace{0.2cm}  
\caption{Few examples of encoder decoder inputs and outputs to the model based on the objectives. Colored words denote special tokens. The text following the \textcolor{blue}{$<$context$>$}  section contains the natural language query/text or SQL (whichever applies). The \textcolor{gray}{$<$header$>$} following that consists of the table headers and finally the table content is row-major linearized with the \textcolor{green}{$<$row$>$} tags. The first three rows describe completion and generation objectives. The last row gives an example of a denoising objective. Note that when the input is only tables, then the corresponding linearized input contains the value \textcolor{blue}{$<$context$>$}\textcolor{brown}{$<$missing\_context$>$} in the context section. All our inputs contain tables and the text/SQL part is optional for the datasets which contain only tables and no associated parallel text.}
\label{tab:example_io}
\end{table*}

\section{UniTabPT Framework}
\label{sec:objectives}
UniTabPT uses encoder-decoder architectures, specifically the T5 architecture proposed in \cite{raffel2020exploring}. We slightly abuse the notation "pretraining" in our work to denote intermediate pretraining as we start with the publicly available text checkpoints of the T5 models and continually train them with self-supervised objectives on table data. There are two key points we want to highlight in our pretraining method:

\begin{itemize}
    \item First, we want the model to understand the semantics of SQL queries along with natural language text, but grounded with the tables they are associated with. 
    \item We want the model to specialize on three kinds of inputs: (1) tables only (2) tables along with text and, (3) tables along with SQL. This is intuitively a straightforward choice as we hypothesize that if the model learns how to associate SQL and text to the table content better, then that would translate to the finetuning stage tasks even when they have fewer annotated labeled data.
\end{itemize}

One of the challenges of table pretraining with heterogeneous inputs is the balance of the scale of data present for different input types. In our work, obtaining good SQL and NL questions for tables is a bottleneck. We linearize the table inputs to row major format and use three unsupervised objectives that allow the model to learn table specific properties. We hypothesize that it would help overcome the scarcity of data and allow the model to learn the inductive bias that can transfer to downstream tasks with related but different inputs. Our work is close to the objectives used in \cite{andrejczuk2022table} although we introduce different generation objectives. As will be described in the following section, our dataset is a mix of table only data and table+text/SQL data. 

There are three elements of the table structure that we need for the denoising objectives: (1) the natural language text or the SQL that represents the query (we will often refer to this part as simply \textit{text} in the context of the inputs), (2) the table headers and, (3) the table cells which contain the table values. We unify all the possible input types i.e. tables, tables+text and tables+SQL into the same format described in Table~\ref{tab:example_io} and use the following objectives. More details on the objectives are present in Section 2 of the Appendix. \\

\noindent \textbf{Denoising}: We use the following masking based text to text input/output objectives. \\ \textbf{1. MLM on table cells}: We use the denoising objectives \cite{yin2020tabert,herzig-etal-2020-tapas,yang-etal-2022-tableformer,wang2021tuta,deng2020turl,dong2022table} on table cell content similar to T5 style masking that was adopted in the original T5 model pretraining with text only tokens. \\ \textbf{2. MLM on NL/SQL text}: We use the same masking as the table cell masking mentioned above. The only difference is the larger proportion of tokens we mask in this component of the input. Note that in both these masking techniques, we mask tokens and not entire cells. \\
\textbf{3. MCP on table headers}: We follow the work done in TaBert \cite{yin2020tabert} and add Masked Column Prediction on the table headers. Unlike MLM, we mask all tokens belonging to a single header value. \\

\noindent \textbf{Generation and Completion}:
We define another set of pre-training strategies inspired by ToTTo \cite{parikh2020totto} in addition to denoising. For each table, we apply the following: for 50\% of the examples in our pretraining dataset, we use the table as input and the corresponding text as output which we term \textit{generation} (see row 3 in Table~\ref{tab:example_io}). For the rest 50\%, we split the text into two halves, where the first half is concatenated alongside the serialized table data to be sent as input to UniTabPT and the second half is then used as the target for prediction.  This is similar to Prefix LM and allows the model to reconstruct the context and forces the model to learn associations between the context and the table content. We term this \textit{completion}(see rows 1 and 2 in Table~\ref{sec:objectives}). We apply these two objectives for both text and SQL inputs alongside the table.

\subsection{Adding special tokens to inputs}\label{sec:special_tokens}
Identifier tagging in inputs has been utilized for code pretraining tasks \cite{wang2021codet5} and mostly in the form of special tokens in instruction finetuning where each task can be prompted with a special token \cite{sanh2021multitask}. Since our aim is to use the model for diverse downstream tasks, we use several special tokens which are marked by colors in the examples shown in Table~\ref{tab:example_io}. There are three kinds of special tokens used: (1) the separators which identify the parts of the table and text inputs and (2) the task identifiers like \texttt{$<$NL\_completion$>$} which identify what the objective corresponds to and, (3) the type of the context/text, whether it is natural language text or SQL. We will later show in the experiments section how this helps us in the downstream tasks. Apart from this, we also add several special tokens for missing values. We add these special tokens to the tokenizer so that they can be treated as single tokens (details in Section 5 of Appendix).

\subsection{Combining the objectives} 
To combine the objectives in a multi-task fashion, we follow the T5 style multi-task learning and following the recent trend \cite{tay2023ul2} in mixing objectives during pretraining. For each example in a batch, we toss a fair coin and choose the denoising objectives 60\% of the time and the generation objective rest of the time. When the denoising objective is selected for a particular example, we apply a few heuristics for the model to learn the table structure better considering the three components of the input. These could be described as such: (1) when the input is table only, we consider MLM on table cells and MCP for headers simultaneously with 50\% probability. For the rest of the time, we only apply MCP with headers. The goal of the second step is for the model to recover the column names based on the table input and without any other corruption, (2) when the input is table and surrounding text or SQL, we apply MLM on table cells and text and MCP on headers simultaneously.

\section{Pretraining Data}\label{sec:pretrain_data}

One of the salient aspects of our work is that we attempt at combining table only inputs along with tables+text and tables+SQL inputs in a unified pretraining stage. We hypothesize that models develop table understanding through these cross correlations that can transfer well to supervised downstream tasks during finetuning. To that end, we collect the following datasets which have mostly been used in most of the previous studies on tabular model pretraining so far, yet studied individually across papers. Note that these datasets contain a mix of table+text (like metadata, captions, questions), table + SQL and table only examples. Since the amount of manually filtered and good quality annotated  parallel table+text and table+SQL data is small compared to all the available data, we use some datasets where we collect the surrounding text like captions and page title of the page they were crawled from and use them as the parallel text. A detailed description of each dataset and statistics can be found in Section 3 of Appendix. \\
\noindent \textbf{1. TAPAS}: We use the data used in the TAPAS paper \cite{herzig-etal-2020-tapas} to collect the data for Wikitables and Infobox. \\
\noindent \textbf{2. Dresden Web Tables}: The Dresden Web Tables Corpus \cite{Eberius:2015} is a collection of about 125 million data tables extracted from the Common Crawl. \\
\noindent \textbf{3. TAPEX}: w use the dataset used in training the TAPEX model \cite{liu2021tapex} and the data is a synthetic corpus of tabular data along with synthetic SQL. \\
\noindent \textbf{4. ToTTo}: We use the data The dataset used in the ToTTo paper \cite{parikh2020totto} which is a collection of 121,000 English Wikipedia tables with corresponding natural language descriptions. \\ 
\noindent \textbf{5. Tabert}: The dataset used in training the tabert model \cite{yin2020tabert} is a collection of 26 million web tables and their associated natural language context. \\
\noindent \textbf{6. GAP}: We use the data used in \cite{shi2022generation} which synthesizes the training data by collecting parallel sentences for the tables. Aggregating all these data, we have roughly 50 billion tokens to train the models on. \\
\begin{table*}[t!]
\centering
{\small
\begin{tabular}{c@{\hspace{0.2cm}}|c@{\hspace{0.6cm}}c@{\hspace{0.6cm}}c@{\hspace{0.6cm}}c@{\hspace{0.6cm}}c@{\hspace{0.6cm}}c@{\hspace{0.6cm}}c@{\hspace{0.1cm}}}  
\toprule \addlinespace[1.2ex]
 \Large \textbf{Model (Parameters)} & \Large \textbf{WikiTQ} & \Large \textbf{WikiSQL} & \Large \textbf{Tabfact} & \Large \textbf{SQA} & \Large \textbf{CoSQL} & \Large \textbf{Sparc} & \Large \textbf{FetaQA} \\  \toprule \addlinespace[1.2ex]
\textbf{BART (400M) }   & 27.93  & 81.62 & 79.04   & - & -   & - & 27.11  \\ 
\textbf{TAPEX (400M) }   & 57.41   & 87.93 & 83.13  & - & -  & - & 25.23 \\ 
\textbf{REASTAP (400M)}   & 57.93   & 87.16 & 83.28   & 56.41 & 41.23   & 58.91 & 26.41  \\ \toprule \addlinespace[1.2ex]
\textbf{T5-large (770M)}   & 43.8   & 86.88 & 80.48  & 55.31 & 49.10  & 56.77 & 29.19  \\ 
\textbf{T5-3B  (3B) } & 50.6   & 89.13 & 85.52   & 59.23 & 54.33   & 63.09 & 30.91  \\
\textbf{Flan-T5-Large (770M)  }  & 45.4   & 87.1 & 82.84   & 58.16 & 56.22  & 61.92 & 31.86   \\
\textbf{Flan-T5-XL (3B) }  & 53.57   & 89.91 & 87.62  & 67.81 & 57.39  & 63.30 & 32.16  \\ \toprule \addlinespace[1.2ex]
\textbf{T5-large + UniTabPT }  & 44.79   & 87.24 & 83.41   & 57.15 & 49.26   & 57.17 & 30.27  \\ 
\textbf{T5-3B  + UniTabPT}  & 53.76   & 89.26 & 86.83   & 65.72 & 55.16   & \textbf{64.82} & 32.14 \\
\textbf{Flan-T5-Large + UniTabPT} & 50.82   & 87.32   & 83.41 & 63.86  & 57.73 & 62.79 & 31.98  \\
\textbf{Flan-T5-XL + UniTabPT } & \textbf{60.29}   & \textbf{90.1} &  \textbf{89.12}   &  \textbf{70.86} &  \textbf{58.71}   &  64.71 &  \textbf{33.12}  \\ \toprule \addlinespace[1.2ex]
\textbf{Flan-T5-XL + UniTabPT + RF}  & 57.96  & 89.92 &  88.71   &  69.27 &  58.49   &  63.81 &  33.08  \\
\end{tabular}
}
\vspace{0.2cm}  
\caption{Results showing the performance of the models on the dev set of the structured knowledge grounded tasks. - denotes we did not run the models on these tasks as we mainly wanted to compare the BART and TAPEX models on specific tasks. Higher the better for these metrics. For the database tasks namely CoSQL and Sparc, we report the average over Exact Match and Execution metrics. For the rest of the results, we report the corresponding metrics as reported in Table 4 of UnifiedSKG \protect \cite{xie2022unifiedskg} For WikiSQL, we report the results for the fully-supervised setting. Bold values denote the best performance in the single task finetuning setting.} 
\label{tab:aggregate_uskg}
\end{table*}
We did a few data processing steps for each of the datasets, but we mainly followed the preprocessing steps mentioned in the paper which produced the Tabert model \cite{yin2020tabert}, and in the interest of the paper length, we list the steps for table processing in Section 4 of Appendix.

\section{Evaluating UniTabPT} \label{sec:finetune}

To answer the three questions we laid out in the Introduction, we evaluate the pretrained model through single task finetuning on several table tasks of generation and classification. We apply  UniTabPT pretraining on the public text checkpoints of 770M and the 3B versions of the T5 and the Flan-T5 models although we later show comparison of our models on the 11B model versions. All details of the pretraining configurations, hyper-parameters and the training resources used have been detailed in Section 5 of the Appendix.
In all of our experiments, we use the same input format for the downstream finetuning as we did for the pretraining mentioned in Table~\ref{tab:example_io}. We only use the encoder input formats and the relevant special tokens during downstream task finetuning and do not use the special tokens for the decoder side inputs. One additional heuristic we use for specializing our model to multi-turn dialogues on tables is that for examples, where we had multiple queries associated with the tables or examples which had both query and metadata along with tables in the pretraining data, we aggregated them in the following format: \texttt{$ \text{query} \ || \ \text{query 2/metadata} \  | \\ \text{query 3/metadata} \  | \ \ldots$ \ | \text{table data } \ldots}. We show how this helps later in downstream tasks with multi-turn dialogues/QA on tables.

\subsection{Structured Knowledge Grounded Generation Tasks}
The goal of the work done in UnifiedSKG \cite{xie2022unifiedskg} was to demonstrate that a single unified text-text framework is able to achieve (near) state-of-the-art performance on all structured knowledge grounded tasks, using a single, general-purpose approach. As mentioned by the authors, their work did not focus on a pretraining approach, rather on a novel methodology to unify different knowledge grounded tasks with a structured text-text framework suitable for sequence to sequence models. We reuse their framework\footnote{\url{https://unifiedskg.com/introduction/#what-is-structured-knowledge-grounding}} for our evaluation purposes with single task finetuning. We use the following tasks for our paper: WikiSQL, WikiTQ, CoSQL, Sparc, FetaQA, SQA and TabFact. We refer the audience to the paper \cite{xie2022unifiedskg} for more details on these tasks and Section 7 of the Appendix for details on these finetuning tasks. We use the train splits of the datasets as used in this framework \footnote{\url{https://github.com/HKUNLP/UnifiedSKG}} for training and the dev set for the evaluation metrics. We do not use the test sets to report the results.

Among these tasks, Sparc and CoSQL are database related tasks, and database schema can widely differ from the schema of tables, so we want to check whether the knowledge gained from table related data can transfer to these domains. We choose these seven tasks among all the tasks mentioned in  UnifiedSKG, since we do not want to evaluate our model on tasks that use knowledge graphs or triples like table + text + passage as inputs. For each task, we report the results using our own finetuning and hyper-parameter search, and in the process, we were not able to match all the metrics exactly as reported in \cite{xie2022unifiedskg} most likely due to differences in hyper-parameter settings including the number of epochs to convergence. We list all the hyper-parameters/configurations for finetuning UniTabPT and the baseline models on each task separately in Section 6 of the Appendix. We consider three baselines specialized on table data for our work: TAPEX which specializes on QA with tables \cite{liu2021tapex}, BART which is the base model TAPEX uses \cite{lewis2019bart} and REASTAP \cite{zhao2022reastap} which is pretrained on synthetic data also starting with BART pretrained checkpoint. We use these baselines since REASTAP has shown to outperform other table pretrained models specialized for these tasks like TAPAS \cite{herzig-etal-2020-tapas}, GraPPa \cite{yu2020grappa}, Tableformer \cite{yang-etal-2022-tableformer}, TaBERT \cite{yin2020tabert} and so these studies are subsumed for the comparisons. All these models are 400M parameter variants and at the time of our study, we did not find larger model sizes from these studies for our comparison. We also note an implicit challenge of finding one baseline model that could be used for all the above tasks to compare with our model. \\

\noindent \textbf{For answering RQ1} with generation tasks, we first evaluate whether our pretraining helps in improving the performance of the models on the tasks individually. We report the single task finetuning results in Table~\ref{tab:aggregate_uskg}. Note that in all the models without UniTabPT pretraining, we do not modify the input structure during finetuning compared to the setting used in UnifiedSKG and the tables are serialized in a row major format. We separate the comparisons in three phases:

\begin{itemize}
    \item \textbf{Comparison with baselines}: We observe that both TAPEX and REASTAP 400M model variants perform much better than even the vanilla 3B parameter T5 and Flan-T5 models  on WikiTQ and WikiSQL tasks which shows that the nature of synthetic data used in these models with reasoning objectives is helpful. However, for other tasks especially, for multi-turn dialogue table tasks such as SQA and CoSQL, we find that REASTAP performs worse than the 770M and 3B Flan T5 variants.  Furthermore, we also conclude that while models like TAPEX and REASTAP are specialized to some tasks due to the nature of the synthetic data they use for the pretraining, our models with UniTabPT can overcome that and show consistent performance gains for all tasks. As one data point, we observe that our Flan-T5+UniTabPT(770M) model is almost 40\% better than REASTAP (400M) model on CoSQL while they are still comparable on WikiTQ (on which the baselines are specialized) and where they have slightly better performance than our model.
    \item \textbf{Comparing the T5 model families}: We observe that the instruction finetuned versions of the T5 models, namely the Flan model families, perform better than their T5 base models on all tasks despite the fact that we do not use in-context learning with instructions, rather the full finetuning setting. We attribute this performance improvement to the fact that the instruction finetuning data in Flan models contain several QA and reasoning tasks. When we apply UniTabPT on top of the T5/Flan T5 models, we observe that the intermediate pretraining improves the performance of the models on all tasks. The gains are almost linear and consistent considering the performances of the baseline T5 and Flan T5 variants. We also observe that for tasks such as the WikiTQ and SQA, the Flan T5 large (770M) model along with UniTabPT is comparable to the T5 3B model for most tasks and in some cases, better. This goes to show that such pretraining can stretch the limits of smaller models for table tasks and can mitigate the latency overhead that comes with larger models with comparable performances. Aggregating these, we conclude that continually pretraining on large scale data can still bring improvement to the already instruction-finetuned Flan models.
    \item \textbf{Comparisons on database and multi-turn dialogue tasks}: As one of the questions we set out to explore, we find that for tasks like CoSQL and Sparc which have database schema as inputs, our pretraining paradigm improves the performance of these models despite that our pretraining did not account for database inputs specifically in the form of database schema inputs. When we compare the improvements we see in SQA which is a form of multi-turn QA on tables, we see that UniTabPT brings significant improvement and we attribute that to the form of multi-turn context we curate for some of our inputs during pretraining.
\end{itemize}

\begin{table}[!t]
\centering
{\small
\begin{tabular}{c@{\hspace{0.6cm}}c@{\hspace{.4cm}}|c@{\hspace{.4cm}}c@{\hspace{.2cm}}}  
\hline
\\
 & \parbox[t]{3cm}{\textbf{Flan-T5-XL + $\text{PFT}$}} & \parbox[t]{3cm}{\textbf{Flan-T5-XL + UniTabPT + $\text{PFT}$}} \\ \\ \hline \\
WikiTQ    & 54.21   & \textbf{61.35}  \\ 
WikiSQL    & 89.96    & \textbf{91.13}     \\
Tabfact    & 70.39   & \textbf{78.64}     \\
SQA     &  52.83   & \textbf{60.95}     \\
CoSQL    & 36.74  & \textbf{40.13}   \\
Sparc     &  51.27   & \textbf{56.38}    \\
FetaQA    & 14.21    & \textbf{19.83}   \\ \hline
\end{tabular}
}
\vspace{0.2cm}  
\caption{Results showing the performance in the low data-resource setting. The results are not reported on the single task finetuned setting but with the setting after the prefinetuning (PFT) stage. The results are evaluated in the low data resource setting (finetuned on 30\% of training data) and the PFT stage did not include the exact data of these tasks. }
\label{tab:pft}
\end{table}

\noindent \textbf{For answering RQ2}, we attempt at improving the generalization performance of the model even beyond UniTabPT with multi-task pre-finetuning followed by task specific finetuning on few samples. The goal is to understand whether UniTabPT can enable the models for better cross task transfer with multi-task pre-finetuning and enable better performance in situations where table task data can be scarce. This would enable us to verify that our \texttt{pretraining $\rightarrow$ multi-task prefinetuning} paradigm can still be useful for direct application in table tasks where data can be scarce. In UniTabPT, we did not explore any data sources that can help the model learn the  alignment between the SQL and the text for a table and which we believe can improve the results on table semantic parsing tasks. In light of these observations and the recent positive results shown from pre-finetuning \cite{aribandi2021ext5} due to cross-task transfer learning, we curate a multi-task finetuning dataset using the following: (1) dataset used in the TAPEX paper where the inputs to outputs are of the form (sql,table) $\rightarrow$ text samples, (2) training sets of the data from  WikiTQ, WikiSQL, KVRET tasks in UnifiedSKG suite, (3) text $\rightarrow$ sql pairs from all the training datasets in UnifiedSKG suite and, (4) in order to avoid catastrophic forgetting of the data from the unsupervised stage \cite{chen2020recall}, we replay a part of the training dataset from the generation objective. 

Note that this is a form of multi-task learning where we control the proportion of datasets to mix from among the above datasets from (1) to (4) so as not to overfit to any particular task. We end up having one million samples in this stage combining all the respective portions. The detailed proportion of the datasets (1) to (4) used in this stage is listed in Section 8 of the Appendix. We also add the column types (identified using Spacy) to the linearized inputs in this stage. We finetune the \texttt{Flan-T5-Xl} and \texttt{Flan-T5-Xl+UniTabPT} models on this dataset for 200 epochs. For evaluation,  we directly evaluate the prefinetuned (PFT) model on the dev sets of WikiSQL and WikiTQ without any further finetuning on them. For other tasks, we finetune them with only 30\% of the available training data. From the results in Table~\ref{tab:pft}, we observe that for all the tasks where we perform finetuning in a low-data resource scenario with only 30\% of the training data, the degradation with PFT is much less when we apply UniTabPT on top of Flan-T5-XL. We conclude that in these situations of low training data, transfer learning with data from different tasks as achieved through prefinetuning helps in better generalization in downstream tasks even when the nature of the inputs and outputs can vary.

One of the advantages of using the special tokens and serializing all task inputs to a unified format is that the model retains some of the properties it learnt about tables and parallel text and we run an experiment to measure the impact that having a unified format brings to downstream tasks. To that end, we finetune a model on the SKG tasks where we do not use the special tokens from our unified format in both the encoder and decoder inputs. Additionally, we assign different tokens to the missing values and separate the table components by ",". We name this format \texttt{RF}. The results of the model Flan-XL+UniTabPT+RF in table~\ref{tab:aggregate_uskg} show clearly that removing the serialization format results in degradation of the performance on the tasks and in many cases wipe out the advantage of UniTabPT. 

\begin{table}[!t]
\centering
{\small
\begin{tabular}{c@{\hspace{0.1cm}}c@{\hspace{0.03cm}}c@{\hspace{0.05cm}}c@{\hspace{0.cm}}}  
\hline
\\
 &  \parbox[t]{1.5cm}{\textbf{BERT \\ Large}} &  \parbox[t]{1.5cm}{\textbf{$\text{Tabert} \\ \text{(K=3)}_\text{{Large}}$}} & \parbox[t]{1.5cm}{ \textbf{Flan-T5 Large} + \\  \textbf{UniTabPT Encoder}} \\ \\ \hline \\
WikiTable-CTP    & 86.31   & \textbf{87.67}  & 87.23  \\ 
WikiTable-CRP  & 85.49   & 86.22  & \textbf{86.92}  \\
Viznet-CTP   & 90.26  & 91.59  & \textbf{91.71}  \\ \hline
\end{tabular}
}
\vspace{0.2cm}  
\caption{Results showing the performance of encoder models on table classification tasks. Results denote micro F1 scores.}
\label{tab:result_class}
\end{table}

\subsection{Table Classification Tasks} \label{sec:classification}
We use the tasks proposed in \cite{suhara} where they study the problem of annotating table columns i.e., predicting column types and the relationships between columns using information from the table exclusively and without any surrounding context. The study uses a multi-task learning framework and uses pre-trained language models for the prediction and we replace their pretrained encoders with our model while keeping the rest of the framework same. We use the problem description from that study to describe the prediction tasks: (Column type prediction, CTP) which annotates a given column from among predefined types and (Column relation prediction, CRP) which annotates the relation between a pair of columns from a pre-defined label set of relations. These are standard classification tasks with inputs and labels. Notably, both these tasks require extracting the representations of the tables before using them with additional layers for classifications. We chose this task to study whether our pretraining can be useful for these classification tasks albeit only with the encoder part of the models. 

We extract the encoder part of the \textit{Flan-T5-large+UniTabPT} model and use the mean of the embeddings from the last layer to get an equivalent representation of the [CLS] token in BERT \cite{ni2021sentence}. The goal of this experiment is to measure the effectiveness of our training when using the encoder for non-generative table specific tasks. To that end, we compare our encoder with the BERT large encoder and the TaBert encoder since they are roughly of equal model sizes. Note that the encoder part of our Flan-T5-large+UniTabPT model is roughly equal to these baseline encoder model sizes. We observe results on the WikiTable dataset and the Viznet dataset for column type (CTP) and column relation prediction (CRP) tasks \cite{suhara}.  We use the framework used in Doduo\footnote{https://github.com/megagonlabs/doduo} while keeping the same hyper-parameters in each run. Table~\ref{tab:result_class} shows the F1 results for the Wikitable tasks and Micro F1 for the Viznet task. Note that we use the same linearization format as used during pretraining,so as to take advatnage of the learned knowledge acquired during the tasks. We observe the following: for all the tasks, our encoder model performs better than BERT which confirms that a unified pretraining framework can be useful even for classification tasks. Our model lags behind Tabert on the Wikitable-CTP task and we hypothesize that since we did not use the column types during pre-training, it could be one reason although we do not see the same sensitivity for Viznet-CTP task. \\

\begin{table}[!t]
\centering
{\small
\begin{tabular}{c@{\hspace{0.2cm}}c@{\hspace{.1cm}}c@{\hspace{.1cm}}c@{\hspace{.1cm}}}  
\hline
\\
 & \parbox[t]{1.5cm}{\textbf{T5 11B}} & \parbox[t]{1.5cm}{\textbf{Flan-XXL}} & \parbox[t]{1.8cm}{\textbf{Flan-XXL+\\UniTabPT}} \\ \\ \hline \\
WikiTQ    & 56.32   & 57.89   & \textbf{62.45}  \\ 
WikiSQL    & 90.16   & 90.83  & \textbf{90.86}     \\
Tabfact    & 88.64   & 89.07   & \textbf{90.71}     \\
SQA     &  69.57  & 70.25   & \textbf{71.23}     \\
CoSQL    & 58.17   &  59.21  & \textbf{59.26}   \\
Sparc     &  65.81 &  67.19   & \textbf{67.41}    \\
FetaQA    & 32.95  & 33.91   & \textbf{35.15}   \\ \hline
\end{tabular}
}
\vspace{0.2cm}  
\caption{Results showing the performance when scaling the model sizes to 11B while reducing the number of tokens trained with.}
\label{tab:11b_scaling}
\end{table}

\noindent \textbf{For answering RQ3}, we compare the impact of the pretraining when scaling the model to 11B parameters. We train the model on roughly two-thirds of the number of epochs/tokens used to train the 770M and 3B models. Due to compute budget limitations, we perform the pretraining only on the Flan-T5-XXL (11B parameters) model and we evaluate three models on the downstream SKG tasks: the T5 11B model, the Flan-T5 11B model and our Flan-T5-XXL+UniTabPT. Table~\ref{tab:11b_scaling} shows that even at the 11B model size, we still see improvements coming from the table pretraining albeit with smaller margins. We observe the following: we see gains with our UniTabPT pretraining comparing Flan-XXL and our model across all the tasks, however, the gains are diminishing when comparing the 770M and the 3B model versions. This suggests that even with the limited data we have, we can still keep getting improvements with table specific pretraining at the scale of 11B model sizes. However, we do observe that the improvements for SQL tasks are marginally lower compared to non-SQL tasks, and we attribute that to the smaller SQL pretraining dataset. One way to mitigate that could be to upsample the SQL data in our pretraining mix which we leave as future explorations.

\section{Related Work}

LLMs have become one popular medium to solve table related tasks \cite{parikh2020totto,wang2020understanding,iida-etal-2021-tabbie,eisenschlos2020understanding}.Most of these tasks are encoder only that do not adapt well to generation tasks. Table semantic parsing and table QA have been solved with different styles of pretraining. Table pretraining tasks mainly aim at understanding the structure of tables and extending them to retrieve task-agnostic generalized representations of tables that can be used for different downstream table-based tasks \cite{herzig-etal-2020-tapas,iida-etal-2021-tabbie,deng2020turl,wang2021tuta}. These studies use different deep neural network architectures mainly in the form of transformer encoders that in addition to taking the entire table information in the form of a serialized string,also add different learnable parameters to encode the structure of the tables. Table representation learning that focuses on joint text and table understanding is a field of research that partially overlaps with our work. Another branch of joint text and table understanding work focuses on text generation from tables\cite{parikh2020totto,yoran2021turning,wang-etal-2022-robust,andrejczuk2022table}. In contrast to these studies,our work centers on the importance of unified pretraining and how the scale of data and model impacts the performance. Our work is also closely related to the study done in \cite{giaquinto2023multitask} which studies  large-scale pretraining of encoder-decoder models specialized towards text-to-SQL generation.


\section{Conclusion and Future Work}
We provide a methodical empirical study to evaluate UniTabPT with table data when applied to diverse table tasks. We conclude that despite the current trend of training the models on lot of natural language text tokens including SQL and table data and instruction finetuning them on text QA datasets, there still is a lot of room for improvement when it comes to table generation and table classification tasks. And pre-finetuning can additionally help generalize the models to situations where the task specific data is scarce. We propose a few areas where our work can be extended: the first is the inclusion of synthetic parallel data in the pretraining and/or prefinetuning stage \cite{jiang2022omnitab}. The second is dealing with long context tokens as tables can be large and which can further bring improvements. Major improvements can come from incorporating the structure of the tables inside the model as shown in \cite{andrejczuk2022table}. Apart from that,our experiments show that merely pretraining on tables also brings improvements on database inputs,so one way to extend our work is to focus on pretraining with relational databases and multi-table inputs such that the model can learn the semantics of the tables.

\bibliographystyle{aaai}
\bibliography{sample}

\clearpage

\twocolumn[ 
   \centering
   {\LARGE \textbf{Appendix: Testing the Limits of Unified Sequence to Sequence LLM Pretraining on Diverse Table Data Tasks}} 
   \vspace{1cm} 
   \addcontentsline{toc}{section}{Appendices}
]
\section{Section 1: Unified LLM Pretraining For Tables} \label{sec:unified_1}
We lay out some fundamental use-cases which primarily motivate the need to consolidate the pretraining phase of several tabular tasks. The fragmented nature of tabular data tasks can be roughly categorized into the following:

\begin{enumerate}
    \item \textbf{Generating SQL-like expressions with natural language queries}: The user's query for a database is expressed in the form of natural language text and the goal of the model is to convert the query text into a formal SQL-like expression that can be executed on the database to output the final answer \cite{sun2018semantic,lin2020bridging}.
    \item \textbf{Question Answering on tables}: The user expresses a query as natural language text as before, but instead of an intermediate step to execute the query to fetch the result, the goal of the model is to directly fetch the answer from the table either through aggregate operations or single cell selections \cite{herzig-etal-2020-tapas,chen2020open}.
    \item \textbf{Fetching representations of table elements for classification}: The goal here is to jointly learn contextual representations for utterances and the structured schema of database tables \cite{yin2020tabert} or in many cases extracting column or row representations for use in downstream tasks \cite{fan2022semantics}. These models typically involve encoders and we later show how extract the representations for these cases. These representations can be extracted at different granular levels and are also used for semantic search for table retrieval \cite{wang2021tuta}.
    \item \textbf{Data to Text Generation}: One of the other use-cases of table language models is text generation. It aims to study the problem of natural language generation from tables with logical inference as intermediate steps \cite{chen2020logical,pietruszka2022stable}.
 
\end{enumerate}

\section{Section 2: Pretraining objectives}
\subsection{Denoising}
\noindent \textbf{MLM on table cells}: We use the regular denoising objectives on table cell content similar to T5 style masking that was adopted in the original T5 model pretraining with text only tokens. This is achieved by randomly masking out spans of tokens and training the model to predict a target sequence containing the missing or corrupted tokens in the input table. The target consists of all of the dropped-out spans of tokens, delimited by sentinel tokens. We replace 15\% of table cell tokens  in the input with a mask/sentinel token. This helps the model capture relationships between the neighbouring cells and the related text and table headers. We use a mean span length of 3 for the T5 style span corruption.\\

\noindent \textbf{MLM on NL/SQL text}: We use the same masking as the table cell masking mentioned above. The only difference is the amount of tokens we mask in this part which is 50\%. We use the same mean span length of 3 for the T5 style span corruption. Note that in both these masking techniques, we mask tokens and not entire cells. \\

\noindent \textbf{MCP on table headers}: We follow the work done in TaBERT \cite{yin2020tabert} and add MCP on the table headers. We mask 40\% of the total number of table headers. Unlike MLM, we mask all tokens belonging to a single header value. Intuitively, MCP encourages the model to recover column information from its contexts and in the process learn the relation between table column values and its headers. 

\subsection{Generation and Completion}
For each table, we use the natural text generation abilities of T5 sequence-sequence models and apply the following: for 50\% of the examples in our pretraining dataset, we use the table as input and the corresponding text as output which we term \textit{generation}. For the rest 50\%, we split the text into two halves, where the first half is concatenated alongside the table data to be sent as input to UniTabPT and the second half is then used as the target for prediction in the seq2seq generation objective.  This is similar to Prefix LM and allows the model to reconstruct the context and forces the model to learn associations between the context and the table content. We term this \textit{completion}. We apply these two objectives for both text and SQL inputs corresponding to the table.

\section{Section 3: Pretraining Data}
\begin{table}[!t]
\centering
{\small
\begin{tabular}{c@{\hspace{0.1cm}}c@{\hspace{.1cm}}c@{\hspace{.2cm}}c@{\hspace{.2cm}}}  
\hline
 \rule{0pt}{3ex}  \textbf{Dataset} & \parbox[t]{1.8cm}{\textbf{Category}} & \parbox[t]{1.5cm}{\textbf{Size \\ after \\ filtering}} & \parbox[t]{1.5cm}{\textbf{Avg. \\ \# rows / \\ \# columns}} \\  \\ \hline \\
\parbox[t]{1.8cm}{\textbf{(TAPAS) \\ Infobox/ \\ Wikitables}}    & \parbox[t]{1.8cm}{(T,Text)}   & 5.5M   &    3.6/7.4  \\ \\
\parbox[t]{1.8cm}{\textbf{Dresden Web Tables 2014}}  & \parbox[t]{1.8cm}{T}   & 15M   & 18.3/5.5    \\ \\
\textbf{TAPEX}    & \parbox[t]{1.8cm}{(T, SQL)}    & 3.1M   & 15.3/5.7   \\ \\
\textbf{GAP}    & \parbox[t]{1.8cm}{(T,Text)}    & 2M   & 12.8/5.8    \\ \\
\textbf{Web Tables 2012}    & \parbox[t]{1.8cm}{(T,Text) / T}  & 25M   & 15.4/5.3   \\ \\
\textbf{ToTTo}    & \parbox[t]{1.8cm}{(T,Text)}   & 107K   & 30.3/6.6    \\ \hline
\end{tabular}
}
\vspace{0.2cm}  
\caption{Statistics/properties of the pretraining dataset. \textit{T} in the category denotes a table.}
\label{tab:pft_data_stats}
\end{table}

\noindent \textbf{TAPAS}: We use the data used in the TAPAS paper \cite{herzig-etal-2020-tapas} to collect the data for Wikitables and Infobox. It contains 6.2M tables (3.3M of classInfobox3and 2.9M of class WikiTable).  We also extract related captions and the title from the tables and use them as the corresponding parallel text. \\

\noindent \textbf{Dresden Web Tables}: The Dresden Web Tables Corpus \cite{Eberius:2015} is a collection of about 125 million data tables extracted from the Common Crawl. DWTC is a large and diverse corpus,containing tables from a wide variety of domains,including,Business,Education,Government,Healthcare. For some Dresden tables,we only have the tables and for the rest,we use the captions and titles as the parallel text. \\

\noindent \textbf{TAPEX}: For training on table and SQL data,we use the dataset used in training the TAPEX model \cite{liu2021tapex} and the data is a synthetic corpus of tabular data along with synthetic SQL. It was created by automatically synthesizing executable SQL queries and their execution outputs. The corpus contains over 3M examples,each of which is associated with a SQL query and its execution output. We do not use the execution input in this part although we use that in the prefinetuning part as well. \\ 

\noindent \textbf{ToTTo}: We use the data The dataset used in the ToTTo paper \cite{parikh2020totto} which is a collection of 121,000 English Wikipedia tables with corresponding natural language descriptions.  Each table in the dataset is associated with a natural language description which we use as parallel text. \\

\noindent \textbf{TaBERT}: The dataset used in training the TaBERT model \cite{yin2020tabert} is a collection of 26 million web tables and their associated natural language context. The tables were extracted from the Common Crawl although the timeline of the pages of the common crawl used in the collection does not overlap with the Dresden web tables corpus. \\

\noindent \textbf{GAP}: We use the data used in \cite{shi2022generation} which synthesizes the training data by collecting parallel sentences for the tables. \\

Note that since the Dresden Web tables corpus and the TaBERT training corpus contain examples which are orders of magnitude larger than the other data sources,we downsample data from these two sources while keeping the rest of the data sources as they are. Final numbers are reported in Table~\ref{tab:pft_data_stats}.

\section{Section 4: Pretraining Data Processing}
We did a few data processing steps for each of the datasets, but we mainly followed the preprocessing steps mentioned in the paper which produced the TaBERT model \cite{yin2020tabert}, and in the interest of the paper, we list only some steps for table processing.  . Note that since our downstream tasks do not contain tables with large rows, we did not handle large table pretraining in this paper which we leave as future work. Instead we resorted to table truncation using the following steps:
\begin{itemize}
    \item For table-text pairs, we consider 3-gram overlap between the natural language text and the table rows (by expanding them as a concatenated string) and keep the rows with the highest overlap. We take a maximum among of 40 rows from the overlapping set. If number of overlapping rows are less than 40, we randomly pick from the non-overlapping rows to fill the rest.
    \item For table only examples, we randomly take 40 rows.
    \item We truncate cells within tables including column names to maximum of 10 words. (words are for  space separated tokens). 
    \item We truncate the natural language text or SQL query to 40 words.
    
\end{itemize}
Apart from the above, we follow some standard steps like sanitizing cell text, sanitizing context,  merging similar columns, removing columns with duplicate names, removing columns with invalid headers and removing rows in which all cells have the same text.

\begin{table*}[!t]
\centering
\begin{tabular}{ccccc}
\toprule
 & Batch Size (Samples) & Dropout &  Learning Rate & No. of epochs \\
\midrule
770M & 4096 & 0.1 & 5e-5 & 7 \\
3B  & 4096 & 0.1 & 5e-5 & 7\\
11B & 2048 & 0.15 & 1e-5 & 5 \\
\bottomrule
\end{tabular}
\caption{Hyper-parameter settings in the table pretraining stage. They are kept the same for both the original T5 and the Flan T5 variants for the respective sizes.}
\label{tab:pt_hparam}
\end{table*}

\section{Section 5: Pretraining Details}
For all our models, we use the T5 and Flan T5 architectures of the public models. We refer the reader to the Huggingface APIs we used to download the model during training \footnote{The HF models for T5: \url{https://huggingface.co/docs/transformers/model_doc/t5} and Flan T5: \url{https://huggingface.co/docs/transformers/model_doc/flan-t5}} for the details. For training the models on the collected pretraining tabular dataset with self supervised objectives, we used the same hyper-parameters for all models including the Flan model families ranging from 770M to 11B and we only modify the dropout and the learning rates at the start of the training.  We use an input sequence length of 1024 tokens for all our pretraining runs and we pad and truncate with the tokenizer accordingly. Since our objectives in this stage are self-supervised, the output sequence lengths of the models are determined based on the masking ratio. However, we follow the implementation used in \cite{raffel2020exploring} that first estimates the number of tokens to be masked starting from a pre-determined output sequence length which in our case is also set to 1024 and the masking ratio. Note that we do not pack multiple sequences into the same input but instead use padding tokens when the inputs do not fill 1024 tokens.  For the 770M and 3B models, we set the learning rate to 1e-5 and the dropout to 0.1 as we find that these avoid early over-fitting when starting from the text pretrained checkpoints of the models we considered in this study. Table~\ref{tab:pt_hparam} lists some of the hyper-parameters used in the training. We reiterate that for all our training runs, we start with the publicly available text checkpoints of the T5 and Flan-T5 model families without changing the architectures.  We use AdamW optimizer with $\beta_1, \beta_2$ and $\epsilon$ set to 0.9, 0.95 and 1e-8 respectively. We use a weight decay of 0.1 and gradient clipping of 1.0 and we use a warmup of 1000 steps for all our training runs. We use Deepspeed Zero Stage 2 for distributed training of the 770M and 3B model variants and we use Zero Stage 3 configuration for the 11B models. All our training configurations use bfloat16 as the precision. As mentioned in the paper, we add special tokens to the public tokenizers (for the respective models) so that they are not tokenized during the training procedure - \texttt{<missing\_cell>, <missing\_column>, <missing\_context>}, for missing table cells, table headers and surrounding context and  \texttt{<row>, <header>, \_|, <text\_NL>, <text\_SQL>, <SQL\_completion>, <NL\_completion>} as identifier tags. We especially find that adding the special tokens to the decoder side inputs as shown in  helps in faster convergence of the loss functions especially, when the SQL+table data is smaller conpared to other forms of data used in pretraining.

\section{Section 6: Finetuning Details}
For the evaluation tasks for finetuning, we follow the setup used in UnifiedSKG \footnote{https://github.com/HKUNLP/UnifiedSKG/tree/main} and directly use their code with some modifications to hyper-parameters to get our results. The discrepancies in the results in our paper and the metrics reported in \cite{xie2022unifiedskg} is mainly due to the number of epochs run and the hyper-parameter settings which we found to vary across the tasks for the best results (we could not find the exact settings from the Github repository or the paper). We especially find that the results are sensitive to the hyper-parameters and for purposes for reproducibility, we have listed the main hyper-parameters we tested using grid-search in Table~\ref{tab:ft_hparam}. We serialize the inputs of all the tasks according to the format demonstrated in Table 1 of our main paper. For all our experiments on REASTAP (\cite{zhao2022reastap}) 770M and 3B models, we use Deepspeed Stage 2 with 8 A100 GPUs for the distributed training. For the 11B models, we use Deepspeed Stage 3 with 8 A100 GPUs for the training.

\section{Section 7: Finetuning Tasks}
\begin{table}[!ht]
\centering
{\small
\begin{tabular}{c@{\hspace{0.2cm}}c@{\hspace{.2cm}}c@{\hspace{.3cm}}c@{\hspace{.2cm}}c@{\hspace{.1cm}}}  
\hline
 \rule{0pt}{3ex} \textbf{Task} & \parbox[t]{1.8cm}{\textbf{Category} \\ \textbf{Input} $\rightarrow$ \\  \textbf{Output}} & \parbox[t]{1.7cm}{\textbf{Train \\ Examples}} & \parbox[t]{1.4cm}{\textbf{Input \\ Length (Tokens)}}  \\   \\ \hline \\
\textbf{WikiTQ}    & \parbox[t]{1.8cm}{(Question,\\ T) $\rightarrow$  Answer}   & 11321   &  1024 \\ \\
\textbf{WikiSQL}    & \parbox[t]{1.8cm}{(Text,\\ T) $\rightarrow$ SQL}   & 56355   & 1024     \\ \\
\textbf{Sparc}    & \parbox[t]{1.8cm}{(Multi-Turn,\\ D) $\rightarrow$ SQL}    & 12059   & 512   \\ \\
\textbf{CoSQL}    & \parbox[t]{1.8cm}{(Dialog \\ D) $\rightarrow$ SQL}    & 2164   & 512     \\ \\
\textbf{SQA}    & \parbox[t]{1.8cm}{(Multi-Turn,\\ T) $\rightarrow$ Answer}  & 12275   & 1024  \\ \\
\textbf{FetaQA}    & \parbox[t]{1.8cm}{(Question,\\ T) $\rightarrow$ \\ Free-form \\ Answer}   & 7326   & 512   \\ \\
\textbf{Tabfact}    & \parbox[t]{1.8cm}{(Statement,\\ T) $\rightarrow$ Answer}   & 1024   & 1024    \\ \\ \hline
\end{tabular}
}
\vspace{0.2cm}  
\caption{Statistics/properties of the finetuning SKG datasets. \textit{T} and \textit{D} in the category denotes a table and a database respectively.}
\label{tab:skg_data}
\end{table}

\begin{table*}[!t]
\centering
\begin{tabular}{ccc}
\toprule
 Dataset & Input $\rightarrow$ Output & Proportion of Dataset  \\
\midrule
TAPEX (\cite{liu2021tapex}  & (Text, Table) $\rightarrow$ SQL & 100\% \\
ToTTo \cite{parikh2020totto}  & Table $\rightarrow$ Text & 100\% \\
WikiTQ \cite{pasupat2015compositional} & (Text, Table) $\rightarrow$ Text (Answer) & 100\%  \\
WikiSQL \cite{zhong2017seq2sql} & (SQL, Table) $\rightarrow$ Text (Answer) & 100\%  \\
Pretraining Data & (Text/SQL, Table) $\rightarrow$ Text & 8\%  \\
Spider \cite{yu2018spider} & Text $\rightarrow$ SQL & 100\%  \\
MIMICSQL \cite{wang2020text} & Text $\rightarrow$ SQL & 150\%  \\
SEDE \cite{hazoom2021text} & Text $\rightarrow$ SQL & 150\%  \\
Sparc \cite{yu2019sparc} & Text $\rightarrow$ SQL & 150\%  \\
CoSQL \cite{yu2019cosql} & Text $\rightarrow$ SQL & 150\%  \\
KVRET \cite{eric2017key} & Text $\rightarrow$ SQL & 150\%  \\
FetaQA \cite{nan2022fetaqa} & Text $\rightarrow$ SQL & 150\%  \\
SQA \cite{iyyer2017search} & Text $\rightarrow$ SQL & 150\%  \\
\bottomrule
\end{tabular}
\caption{Datasets used in the prefinetuning stage. Proportion of dataset indicates the proportion of data from the total available samples within that data. When proportion is greater than 100\%, it means we upsample the data by repeating a portion of the dataset again. For datasets which are part of the evaluation tasks in our paper namely, WikiTQ, WikiSQL, Sparc, CoSQL, SQA and FetaQA, we use the portion of the training datasets only, for the partitions we have created prior to evaluation.}
\label{tab:pft_data}
\end{table*}
The goal of the work done in \cite{xie2022unifiedskg} was to demonstrate that a single unified text-text framework is able to achieve (near) state-of-the-art performance on all structured knowledge grounded tasks, using a single, general-purpose approach. As mentioned by the authors, their work did not focus on a pretraining approach, rather on a novel methodology to unify different knowledge grounded tasks with a structured text-text framework. We reuse their knowledge grounding framework for our evaluation with the T5 models. Although their work focused on multiple techniques of finetuning like single task, multi-task, prefix LM among others, in our work, we focus on single task finetuning, as we are mainly interested in measuring the improvements that come from the pretraining stage on different tasks. We use the following tasks for our paper: WikiSQL, WikiTQ, CoSQL, Sparc, FetaQA, SQA and TabFact. We refer the audience to the paper \cite{xie2022unifiedskg} for more details on these tasks. We provide a brief summary of the kind of tasks in Table~\ref{tab:skg_data}. We mention some key points about the datasets:

\begin{itemize}
    \item We use the train splits of the datasets as used in this framework \footnote{https://github.com/HKUNLP/UnifiedSKG} for training and the dev set for the evaluation metrics. We do not use the test sets to report the results.
    \item For the database tasks namely CoSQL and Sparc, we report the average over Exact Match and Execution metrics. For the rest of the results, we report the corresponding metrics as reported in Table 4 of UnifiedSKG \cite{xie2022unifiedskg}.
    \item For the WikiSQL task, we report the results for the fully-supervised setting.
\end{itemize}

As can be noted, among these tasks, Sparc and CoSQL are database related tasks, and database schema can widely differ from the schema of tables, so we want to check whether the knowledge gained from table related data can transfer to these domains. 

\section{Section 8: Prefinetuning data Mixtures}
One of the critical bottleneck of any multi-task learning stage which utilizes data from different tasks is the right mixture of datasets \cite{sanh2021multitask} to use that allows for better generalization. To that end of ensuring that the prefinetuning stage enables better generalization in cases where downstream table tasks may have scarcity of training data, we utilized the proportions of the datasets detailed in Table~\ref{tab:pft_data}. Note that all of these are used for sequence to sequence generation objective and without any masking where the inputs to output generation is the objective itself. So this is a supervised learning stage. A few key points to note about the dataset and the way we use it for evaluation of this stage in the main paper. Since we use the WikiSQL and the WikiTQ training splits of the datasets in its entirety and since they also are tasks in our evaluation suite, we perform zero shot evaluation after this stage is applied and report the results in the main paper. For other tasks like CoSQL, SQA and FetaQA which are also part of our evaluation suite, we only use the SQL input and the NL text output parts of these datasets in the prefinetuning data mixture. Note that these are all (Text, Table) $\rightarrow$ SQL datasets and we do not use the table portions. The main reason behind this is that we want the model to learn text and SQL alignment which is one of the characteristics that the model would not learn well in the pretraining stage. Following this, for the evaluations, we only use 30\% of the trainining dataset of each task when we evaluate the prefinteuned models on these tasks, so there will be some overlap between the training dataset and the portion of the dataset used in the prefinetuning stage (although the exact input to output formats are different due to table exclusion). In all of these, there is no leakage of the dev sets of the tasks into the prefinetuning data mixtures.


\begin{table*}[!t]
\centering
\begin{tabular}{c@{\hspace{0.25cm}}c@{\hspace{.25cm}}c@{\hspace{.2cm}}c@{\hspace{.25cm}}c@{\hspace{.25cm}}c@{\hspace{.25cm}}c@{\hspace{.25cm}}c@{\hspace{.25cm}}c@{\hspace{.25cm}}} 
\toprule
 & Task & \parbox[t]{1.8cm}{ Batch Size \\ (Examples)} & Input max length & Dropout &  Learning Rate & Weight decay &  No. of epochs & Beam Size \\
\midrule
770M & WikiSQL  &128 & 1024 & 0.1 & 1e-5 & 0.01 & 60 & 4 \\
3B  & WikiSQL & 128 & 1024 & 0.1 & 5e-5 & 0.01 & 60 & 4\\
11B & WikiSQL & 64 & 1024 &  0.1 & 5e-5 & 0.01 & 60 & 4\\
REASTAP & WikiSQL & 128 & 1024 & 0.1 & 5e-5 & 0.01 & 60 & 4\\ \\
770M & WikiTQ  &128 & 1024 & 0.1 & 1e-5 & 0.01 & 150 & 4 \\
3B  & WikiTQ & 128 & 1024 & 0.1 & 5e-5 & 0.01 & 150 & 4\\
11B & WikiTQ & 64 & 1024 &  0.1 & 5e-5 & 0.01 & 150 & 4\\
REASTAP & WikiTQ & 128 & 1024 & 0.1 & 5e-5 & 0.01 & 150 & 4\\ \\
770M & Tabfact  &128 & 1024 & 0.1 & 1e-5 & 0.1 & 50 & 4 \\
3B  & Tabfact & 128 & 1024 & 0.1 & 5e-5 & 0.1 & 50 & 4\\
11B & Tabfact & 64 & 1024 &  0.1 & 5e-5 & 0.1 & 50 & 4\\
REASTAP & Tabfact & 128 & 1024 & 0.1 & 5e-5 & 0.1 & 50 & 4\\ \\
770M & CoSQL  &128 & 1024 & 0.01 & 1e-5 & 0.01 & 150 & 1 \\
3B  & CoSQL & 128 & 1024 & 0.01 & 5e-5 & 0.01 & 150 & 1\\
11B & CoSQL & 64 & 1024 &  0.01 & 5e-5 & 0.01 & 150 & 1\\
REASTAP & CoSQL & 128 & 1024 & 0.01 & 5e-5 & 0.01 & 150 & 1\\ \\
770M & SQA  &128 & 1024 & 0.1 & 1e-5 & 0.01 & 50 & 4 \\
3B  & SQA & 128 & 1024 & 0.1 & 5e-5 & 0.01 & 50 & 4\\
11B & SQA & 64 & 1024 &  0.1 & 5e-5 & 0.01 & 50 & 4\\
REASTAP & SQA & 128 & 1024 & 0.1 & 5e-5 & 0.01 & 50 & 4\\ \\
770M & Sparc  &128 & 1024 & 0.1 & 1e-5 & 0.01 & 150 & 1 \\
3B  & Sparc & 128 & 1024 & 0.1 & 5e-5 & 0.01 & 150 & 1\\
11B & Sparc & 64 & 1024 &  0.1 & 5e-5 & 0.01 & 150 & 1\\
REASTAP & Sparc & 128 & 1024 & 0.1 & 5e-5 & 0.01 & 150 & 1\\ \\
770M & FetaQA  &128 & 512 & 0.1 & 1e-5 & 0.01 & 150 & 4 \\
3B  & FetaQA & 128 & 512 & 0.1 & 5e-5 & 0.01 & 150 & 4\\
11B & FetaQA & 64 & 512 &  0.1 & 5e-5 & 0.01 & 150 & 4\\
REASTAP & FetaQA & 128 & 512 & 0.1 & 5e-5 & 0.01 & 150 & 4\\ \\
\bottomrule
\end{tabular}
\caption{Hyper-parameter settings for the finetuning tasks. They are kept the same for both the original T5 and the Flan T5 variants for the respective sizes where applicable.}
\label{tab:ft_hparam}
\end{table*}

\end{document}